\documentclass[nohyperref]{article}
\usepackage{float}
\usepackage{graphicx}
\usepackage{hyperref}
\usepackage[utf8]{inputenc}
\usepackage{makecell}
\usepackage{multicol}
\usepackage[multiple]{footmisc}
\usepackage{multirow}
\usepackage{xcolor}
\usepackage{amsmath}
\usepackage[accepted]{icml2022}

\begin{document}

\twocolumn[
\icmltitle{Text and Code Embeddings by Contrastive Pre-Training}

\icmlsetsymbol{equal}{*}

\begin{icmlauthorlist}
\icmlauthor{Arvind Neelakantan}{equal,comp}
\icmlauthor{Tao Xu}{equal,comp}
\icmlauthor{Raul Puri}{comp}
\icmlauthor{Alec Radford}{comp}
\icmlauthor{Jesse Michael Han}{comp}
\icmlauthor{Jerry Tworek}{comp}
\icmlauthor{Qiming Yuan}{comp}
\icmlauthor{Nikolas Tezak}{comp}
\icmlauthor{Jong Wook Kim}{comp}
\icmlauthor{Chris Hallacy}{comp}
\icmlauthor{Johannes Heidecke}{comp}
\icmlauthor{Pranav Shyam}{comp}
\icmlauthor{Boris Power}{comp}
\icmlauthor{Tyna Eloundou Nekoul}{comp}
\icmlauthor{Girish Sastry}{comp}
\icmlauthor{Gretchen Krueger}{comp}
\icmlauthor{David Schnurr}{comp}
\icmlauthor{Felipe Petroski Such}{comp}
\icmlauthor{Kenny Hsu}{comp}
\icmlauthor{Madeleine Thompson}{comp}
\icmlauthor{Tabarak Khan}{comp}
\icmlauthor{Toki Sherbakov}{comp}
\icmlauthor{Joanne Jang}{comp}
\icmlauthor{Peter Welinder}{comp}
\icmlauthor{Lilian Weng}{comp}
\end{icmlauthorlist}

\icmlaffiliation{comp}{OpenAI}

\icmlcorrespondingauthor{Arvind Neelakantan}{arvind@openai.com}

\icmlkeywords{Unsupervised Learning, Embeddings, Contrastive Learning}

\vskip 0.3in
]



\printAffiliationsAndNotice{\icmlEqualContribution} 

\begin{abstract}
Text embeddings are useful features in many applications such as semantic search and computing text similarity. Previous work typically trains models customized for different use cases, varying in dataset choice, training objective and model architecture. In this work, we show that contrastive pre-training on unsupervised data at scale leads to high quality vector representations of text and code. The same unsupervised text embeddings that achieve new state-of-the-art results in linear-probe classification also display impressive semantic search capabilities and sometimes even perform competitively with fine-tuned models. On linear-probe classification accuracy averaging over 7 tasks, our best unsupervised model achieves a relative improvement of 4\% and 1.8\% over previous best unsupervised and supervised text embedding models respectively. The same text embeddings when evaluated on large-scale semantic search attains a relative improvement of 23.4\%, 14.7\%, and 10.6\% over previous best unsupervised methods on MSMARCO, Natural Questions and TriviaQA benchmarks, respectively. Similarly to text embeddings, we train code embedding models on (text, code) pairs, obtaining a 20.8\% relative improvement over prior best work on code search. 
\end{abstract}

\section{Introduction}
\begin{figure}[h]
\centering
\includegraphics[width=0.35\textwidth]{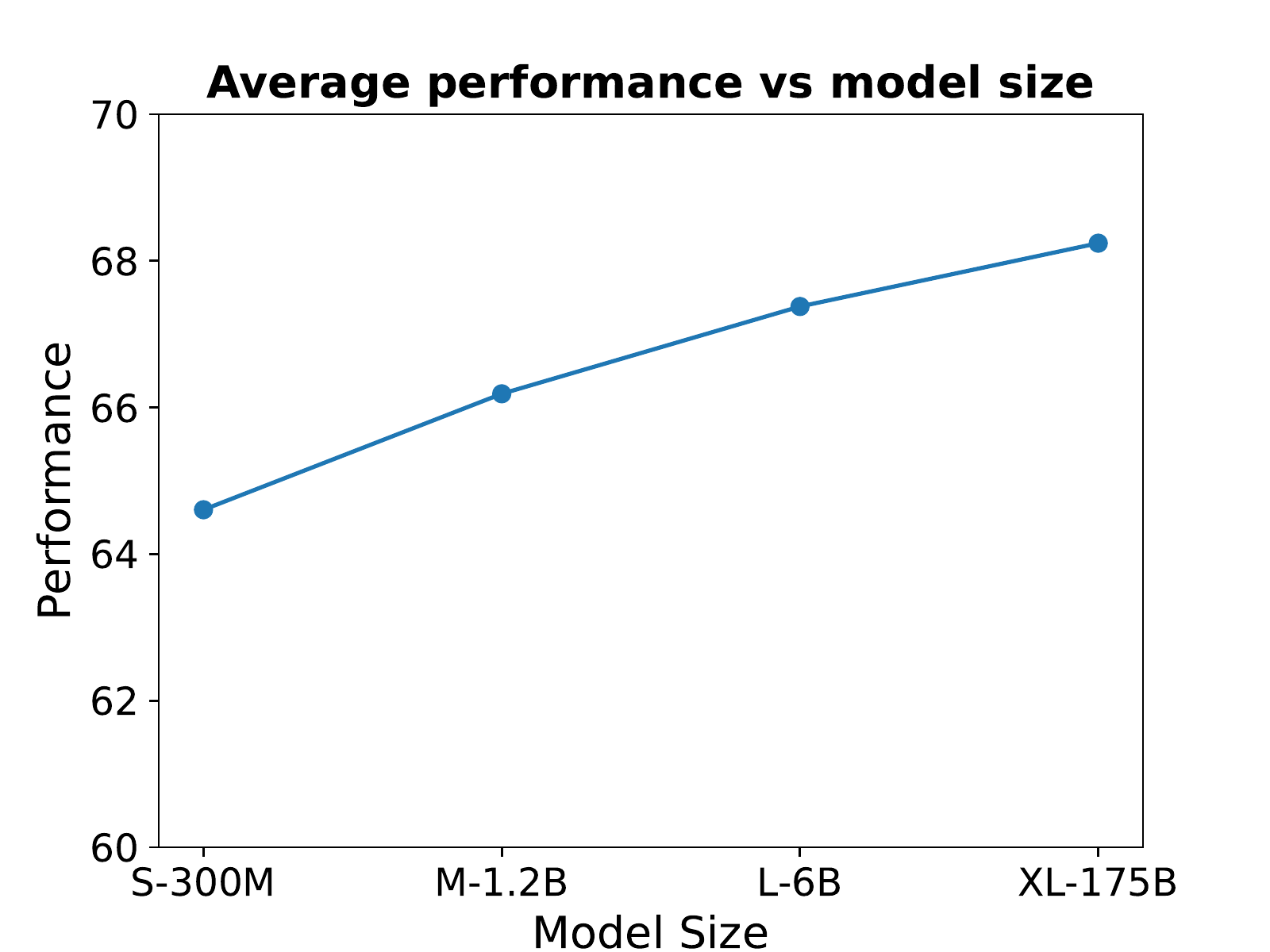}
\caption{Average performance of unsupervised $\texttt{cpt-text}$ models of different sizes across 22 tasks consisting of linear-probe classification, text search, and sentence similarity tasks.}
\label{fig:summary}
\end{figure}
Deep unsupervised learning with generative and embedding models has seen dramatic success in the past few years. Generative models \cite{elmo, t5, wavenet, dalle, gpt-3, codex} are trained to maximize the likelihood of observed data while embedding models are trained to distinguish observed data from noise \cite{inbatch,Oord,clip,align,simcse,contreiver}.  Generative models have been shown to produce realistic content and benefit many downstream applications, reducing the need for labeled training datasets. In generative models, the information about the input is typically distributed over multiple hidden states of the model. While some generative models \cite{vae,Kiros} can learn a single representation of the input, most autoregressive Transformer \cite{transformer} models do not \cite{t5, gpt-3, codex, dalle}. However, learning such a representation (or embedding) is necessary for many tasks. Systems that search over millions or billions of items require each entry to be embedded as a dense representation and build an index in advance to save computational costs at query time. These embeddings are useful features for classification tasks and can also enable data visualization applications via techniques such as clustering. Embedding models are explicitly optimized to learn a low dimensional representation that captures the semantic meaning of the input \cite{clip,align,declutr,simcse,contreiver}.   

In this work, we train embedding models using a contrastive learning objective with in-batch negatives \cite{inbatch,contrastive} on unlabeled data. The input is encoded with a Transformer encoder \cite{transformer} and we leverage naturally occurring paired data to construct training data with no explicit labels. Text embedding models are trained on paired text data where we consider neighboring pieces of text on the Internet as positive pairs. Code embedding models treat the top-level docstring in a function along with its implementation as a (text, code) pair. The training signal of the contrastive objective on its own is not sufficient to learn useful representations and we overcome this by initializing our model with other pre-trained models \cite{gpt-3,codex}. Finally, we find that it is critical to use a sufficiently large batch to achieve the optimal performance. We show that this simple recipe combining pre-trained model initialization, large-batch contrastive learning and training at scale, can produce text and code embeddings that possess a broad range of capabilities.

We train a series of unsupervised text embedding models ($\texttt{cpt-text}$) of different sizes, ranging from 300M to 175B parameters, and observe a consistent performance improvement with increasing model sizes (Figure \ref{fig:summary}). On classification accuracy averaging across 7 linear-probe classification tasks in SentEval \cite{senteval}, our largest unsupervised model achieves new state-of-the-art results with a relative improvement of 4\% and 1.8\% over the previous best unsupervised \cite{declutr} and supervised \cite{simcse} text embedding models, respectively.

Text embedding in previous work was studied under different domains, varying in data, training objective and model architecture. Precisely, sentence embedding \cite{sbert,simcse,declutr} and neural information retrieval \cite{ORQA, REALM, dpr, e2e, contreiver} have remained different research topics evaluated on distinct benchmarks, even though both aim to learn high-quality text representation.
However, we find the same model that achieves good performance on sentence embedding benchmarks, as discussed above, is also able to obtain impressive results on large-scale information retrieval. When evaluated on the MSMARCO passage ranking task \cite{msmarco} to search over 4M passages, $\texttt{cpt-text}$ gets a relative improvement of 23.4\% over previous best unsupervised methods \cite{bm25}.  On the task of searching on 21M documents from Wikipedia, $\texttt{cpt-text}$ obtains a relative improvement of 14.7\%, and 10.6\% over previous unsupervised methods \cite{contreiver} for Natural Questions \cite{nq} and TriviaQA \cite{trivia}, respectively. On TriviaQA, our unsupervised method is even competitive with fine-tuned models. 

Next, we train code embedding models ($\texttt{cpt-code}$) using the same recipe. Our models learn via (text, code) pairs, extracted from open source code. We evaluate our model on CodeSearchNet \cite{codesearchnet}, a commonly used code search benchmark, where the task is to find the most relevant code snippet given a natural language query. Our models achieve new state-of-the-art results with a 20.8\% relative improvement over the previous best result \cite{Guo}. Unlike text embedding models, we observe no performance improvement on code search when increasing the number of parameters of $\texttt{cpt-code}$ from 300M to 1.2B.

Finally, we experiment with fine-tuning our models on several supervised datasets and study the transfer learning performance. When fine-tuned on NLI (Natural Language Inference) datasets, we see a further boost in linear-probe classification, outperforming the previous best transfer method \cite{simcse} by 2.2\%. On SST-2 sentiment classification \cite{sst}, we find that our representations are sufficiently descriptive that even a simple $k$-NN classifier achieves results comparable to a linear-probe classifier. Interestingly, zero-shot performance with our embeddings outperforms the supervised neural network models introduced along with the release of the SST-2 dataset.  We also fine-tune the unsupervised model on MSMARCO and evaluate it on a suite of zero-shot search tasks in the BEIR benchmark \cite{beir}. In the transfer setting, our models achieve a 5.2\% relative improvement over previous methods \cite{contreiver} and is comparable even with methods \cite{colbert,splade,mini} that demand substantially more computation at test time.

\section{Approach}
\label{sec:approach}

Our models are trained with a contrastive objective on paired data. In this section, we present more details on the model architecture and the training objective. The training set consists of paired samples, $\{(x_i,y_i)\}^N_{i=1}$, where $(x_i,y_i)$ corresponds to a positive example pair, indicating that $x_i$ and $y_i$ are semantically similar or contextually relevant.

\subsection{Model}
\label{sec:model}

\begin{figure}[h]
\centering
\includegraphics[width=0.3\textwidth]{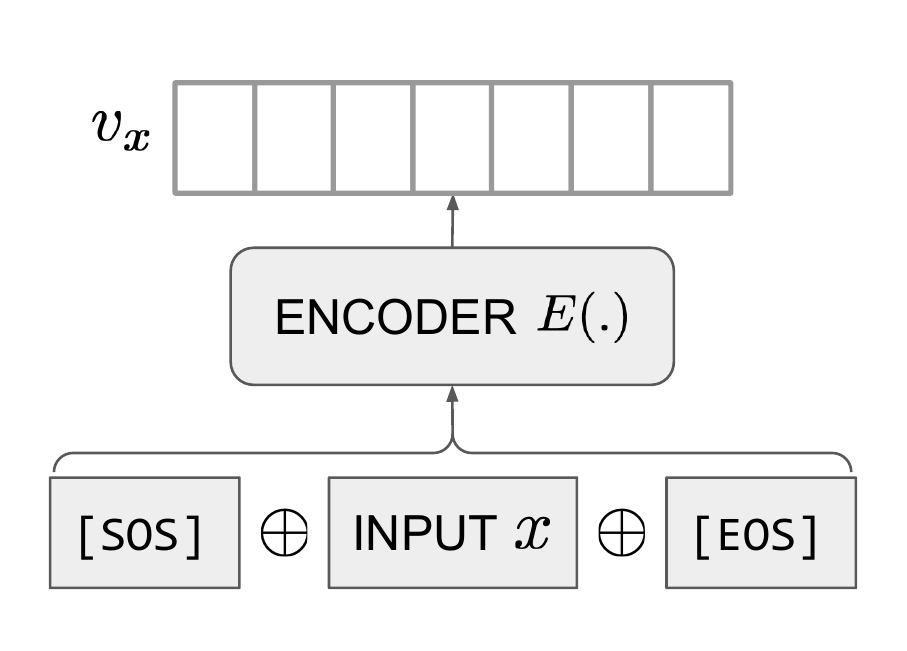}
\caption{The encoder $E$ maps input $x$ to embedding $v_x$. Special tokens, \texttt{[SOS]} and \texttt{[EOS]}, are appended to the start and end of the input sequence respectively. The last layer hidden state corresponding to the token \texttt{[EOS]} is extracted as the embedding of the input sequence.}
\label{fig:model}
\end{figure}

Given a training pair $(x,y)$, a Transformer \cite{transformer} encoder $E$ is used to process $x$ and $y$ independently. The encoder maps the input to a dense vector representation or embedding (Figure \ref{fig:model}). We insert two special token delimiters, \texttt{[SOS]} and \texttt{[EOS]}, to the start and end of the input sequence respectively. The hidden state from the last layer corresponding to the special token \texttt{[EOS]} is considered as the embedding of the input sequence.

\begin{figure}[h]
\centering
\includegraphics[width=0.4\textwidth]{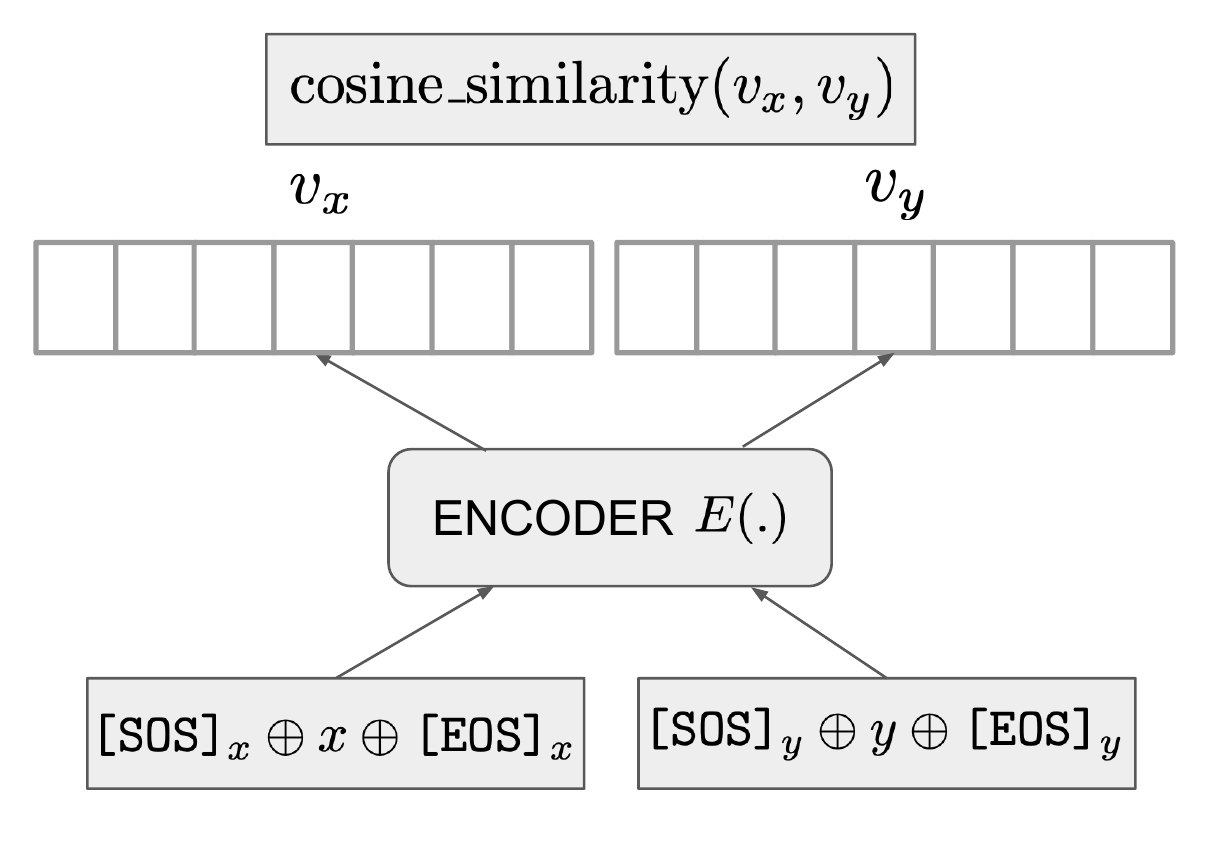}
\caption{The encoder $E$ maps inputs $x$ and $y$, to embeddings, $v_x$ and $v_y$ independently. The similarity score between $x$ and $y$ is defined as the cosine similarity between these two embedding vectors.}
\label{fig:train}
\end{figure}

The Transformer encoder maps the input, $x$ and $y$, to embeddings, $v_x$ and $v_y$ respectively and the similarity between two inputs is quantified by the cosine similarity between their embeddings, $v_x$ and $v_y$ (Figure \ref{fig:train}).

$$
\begin{aligned}
v_x &= E(\texttt{[SOS]}_x \oplus x \oplus \texttt{[EOS]}_x) \\
v_y &= E(\texttt{[SOS]}_y \oplus y \oplus \texttt{[EOS]}_y) \\
\text{sim}(x,y) &= \frac{v_x \cdot v_y}{\|v_x\| \cdot \|v_y\|} 
\end{aligned}
$$
where $\oplus$ is an operation to concatenate two strings together. We found that using different delimiters leads to more stable training. For $x$, we use `['  as $\texttt{[SOS]}_x$ and `]' as $\texttt{[EOS]}_x$, while we use `\{' and `\}' as $\texttt{[SOS]}_y$ and $\texttt{[EOS]}_y$ respectively for $y$. 

\subsection{Training Objective}
\label{sec:training}

The paired samples in the training set are contrasted against in-batch negatives \cite{contrastive,inbatch}. Contrastive learning with in-batch negatives has been widely used for unsupervised representation learning in prior work \cite{clip,align,Chen,contreiver}. For each example in a mini-batch of $M$ examples, the other $(M-1)$ in the batch are used as negative examples. The usage of in-batch negatives enables re-use of computation both in the forward and the backward pass making training highly efficient. The \texttt{logits} for one batch is a $M\times M$ matrix, where each entry $\text{logit}(x_i,y_j)$ is given by,

\begin{equation*}
\begin{split}
\text{logit}(x_i, &y_j)  = \text{sim}(x_i, y_j) \cdot \exp(\tau), \\
 & \forall (i,j), i,j \in \{1,2,\ldots,M\}
\end{split}
\end{equation*}

where $\tau$ is a trainable temperature parameter.

Only entries on the diagonal of the matrix are considered positive examples. The final training loss is the sum of the cross entropy losses on the row and the column direction, as described in the following numpy style pseudo code.


\begin{small}
\begin{verbatim}
labels = np.arange(M)
l_r = cross_entropy(logits, labels, axis=0)
l_c = cross_entropy(logits, labels, axis=1)
loss = (l_r + l_c) / 2
\end{verbatim}
\end{small}

We initialize our models with pre-trained generative language models. $\texttt{cpt-text}$ is initialized with GPT models \cite{gpt-3} and $\texttt{cpt-code}$ is initialized with Codex models \cite{codex}. When fine-tuning our models (Section \ref{sec:results}), the supervised training data like NLI datasets contain explicit negative examples and they are used along with the in-batch negatives. 

\section{Results}
\label{sec:results}

\begin{table}[]
\centering

\setlength{\tabcolsep}{4.5pt}
\begin{tabular}{l|r|r|r}
\Xhline{2.5\arrayrulewidth}
Model & Parameters &  Embed Dimensions & Batch size  \\ 
\Xhline{1\arrayrulewidth} & \\[-1.5ex]
S & 300M  & 1024 & 12288 \\ 
M & 1.2B  & 2048 & 6912 \\ 
L & 6B & 4096  & 5896 \\ 
XL & 175B  & 12288 & 4976 \\ 
\Xhline{2.5\arrayrulewidth}
\end{tabular}
\caption{Batch size used to train the models of different sizes.}
\label{table:bs}
\end{table}

Our models are trained on naturally occurring paired data. $\texttt{cpt-text}$ models are trained on Internet data with neighboring pieces of text as positive pairs for the contrastive objective. The code embedding $\texttt{cpt-code}$ models use (text, code) pairs extracted from open source code. As discussed in Section \ref{sec:bs}, sufficiently large batch size is crucial to achieve good performance with our setup. Table~\ref{table:bs} lists the batch sizes used to train the models of different sizes. 

We evaluate our text embedding models on a broad range of tasks: linear-probe classification, sentence similarity, and semantic search. While sentence embedding \cite{sbert,simcse,declutr} methods report results only on embedding benchmarks and neural information retrieval methods \cite{ORQA, REALM, dpr, e2e, contreiver} report results only on search benchmarks, we use the \textit{same} unsupervised model across all these tasks. 

\begin{table*}
\centering
\begin{tabular}{lcccccccc}
& MR & CR & SUBJ & MPQA & SST & TREC & MRPC & Avg. \\ 
\Xhline{2.5\arrayrulewidth}
\multicolumn{9}{c}{Unsupervised} \\
\Xhline{1\arrayrulewidth} \\[-1.5ex]
BERT \cite{bert} & 78.7 & 86.2 & 94.4 & 88.7 & 84.4 & 92.8 & 69.4 & 84.9 \\
SimCSE \cite{simcse} & 84.7 & 88.6 & 95.4 & 87.5 & 89.5 & 95.0 & 72.4 & 87.6 \\
DECLUTR \cite{declutr} & 85.2 & 90.7 & 95.8 & 88.5 & 90.0 & 93.2 & \textbf{74.6} & 88.3 \\ \hline & \\[-1.5ex] 
$\texttt{cpt-text}$ S & 87.1 & 90.1 & 94.9 & 88.3 & 91.8 & 95.2 & 71.6 & 88.4 \\
$\texttt{cpt-text}$ M & 89.0 & 90.9 & 96.7 & 89.6 & 93.9 & 96.6 & 73.6 & 89.9 \\
$\texttt{cpt-text}$ L & 90.6 & 92.6 & 97.0 & 90.6 & 95.3 & 97.0 & 73.6 & 90.9 \\
$\texttt{cpt-text}$ XL & \textbf{92.2} & \textbf{93.5} & \textbf{97.4} & \textbf{91.5} & \textbf{96.2} & \textbf{97.4} & 74.1 & \textbf{91.8} \\ 
\Xhline{2.5\arrayrulewidth} 
\multicolumn{9}{c}{Transfer from NLI data} \\ 
\Xhline{1\arrayrulewidth} \\[-1.5ex]
SBERT \cite{sbert} & 84.9 & 90.1 & 94.5 & 90.3 & 90.7 & 87.4 & 75.9 & 87.7 \\
SimCSE \cite{simcse} & 88.4 & 92.5 & 95.2 & 90.1 & 93.3 & 93.8 & 77.7 & 90.2 \\ \hline & \\[-1.5ex]
$\texttt{cpt-text}$ S & 87.3 & 91.0 & 94.6 & 90.5 & 91.4 & 95.0 & 75.6 & 89.3 \\
$\texttt{cpt-text}$ M & 89.8 & 92.7 & 95.7 & 91.3 & 95.3 & 96.6 & 76.5 & 91.1 \\
$\texttt{cpt-text}$ L & 90.8 & 93.5 & 96.2 & 91.2 & 95.7 & 96.0 & 76.9 & 91.5 \\
$\texttt{cpt-text}$ XL & \textbf{92.4} & \textbf{93.9} & \textbf{97.0} & \textbf{91.8} & \textbf{95.8} & \textbf{96.4} & \textbf{78.1} & \textbf{92.2} \\
\Xhline{2.5\arrayrulewidth}
\end{tabular}
\caption{$\texttt{cpt-text}$ models of different sizes, ranging from 300M (S) to 175B (XL), are compared to previous work on linear-probe classification tasks in SentEval. We report performance of unsupervised models, as well as those fine-tuned on NLI data.}
\label{table:linear}
\end{table*}

\subsection{Text Embedding}

The SentEval benchmark \cite{senteval} is widely adopted to assess the quality of sentence embeddings, consisting of a broad collection of tasks in the categories of linear-probe classification and sentence similarity, and we use the same to evaluate ours.

\subsubsection{Linear Probe Classification}
\label{sec:linear}

When evaluated on linear-probe classification, the embeddings are used as features to train a linear classifier to solve a variety of downstream tasks. The results in Table \ref{table:linear} demonstrate a clear advantage of larger model sizes producing better features for improved classification performance. In transfer learning setup, we fine-tune unsupervised $\texttt{cpt-text}$ models on SNLI \cite{snli} and MNLI \cite{mnli} datasets using entailment pairs as positive examples and contradiction pairs as negative examples. On both unsupervised learning and transfer learning settings, we achieve state-of-the-art results.

\subsubsection{Zero-shot and $k$-NN Classification}
\label{sec:zs}

In this section, we discuss results using zero-shot classification and $k$-nearest neighbor classification on the SST-2 binary sentiment classification task \cite{sst}. We experiment with 6B (L) $\texttt{cpt-text}$ model fine-tuned on NLI data for this study. In the first zero-shot experiment, each input text is assigned with one of the two labels (`positive', `negative') based on which label has its embedding closest to the input text embedding. The performance can be further improved by prompting, where we use a simple label description, `this is an example of a {positive/negative} movie review.', instead of a single word. This zero-shot usage of embeddings is novel compared to prior work on embeddings and it is interesting to note that our zero-shot results are better than the supervised neural network results reported along with the release of the dataset \cite{sst}. In the $k$-NN classification experiment, given an input text, the prediction is the majority label among 256 training examples closest to the test input in the embedding space. As shown in Table \ref{table:sst}, the $k$-NN classifier without any task-specific tuning of trainable parameters achieves results comparable to a linear classifier.

\begin{table}[]
\centering
\begin{tabular}{l|c}
\Xhline{2.5\arrayrulewidth}
Method                      & Accuracy \\ 
\Xhline{1\arrayrulewidth} \\[-1.5ex]
Zero-shot                   & 88.1 \\
Zero-shot with prompting    & 89.1 \\
$k$-NN                      & 93.3 \\
Linear-probe                & 95.7 \\
Full fine-tuned SOTA        & 97.5 \\
\Xhline{2.5\arrayrulewidth}
\end{tabular}
\caption{Comparison of different classification strategies using the 6B $\texttt{cpt-text}$ model fine-tuned on NLI data for SST-2 binary sentiment task \cite{sst}. Our zero-shot results are better than the $85.4$\% accuracy obtained by supervised neural networks reported along with the release of the dataset \cite{sst}.}
\label{table:sst}
\end{table}

\subsubsection{Sentence Similarity}
\label{sec:sts}

On sentence similarity tasks in SentEval, we find that our models perform worse than previous SOTA methods (Table \ref{table:sts}). Sentence similarity is not a completely well-defined downstream task (e.g. are the sentences, `Jack loves Jill' and `Mary loves chocolates', similar?).\footnote{\url{https://twitter.com/yoavgo/status/1431299645570011142}}\footnote{\url{https://twitter.com/yoavgo/status/1483565266575540225?s=20}} For example, \citet{similarity} argue that two objects can be infinitely similar or dissimilar \cite{Vervaeke2012RelevanceRA}. A possible explanation for why our models perform better than prior work on search and classification but not on these tasks is that our models might not be optimized for the specific definition used by these sentence similarity benchmarks. It is important to note that previous embedding search methods do not report performance on sentence similarity tasks \cite{dpr, e2e, contreiver}.  More discussion on this phenomenon is presented in Section \ref{sec:behavior}.  


\begin{table}[]
\small
\setlength{\tabcolsep}{3pt}
\centering
\begin{tabular}{lcccccc}
\multicolumn{1}{r}{} STS & -12 & -13 & -14 & -15 & -16 & Avg \\ 
\Xhline{2.5\arrayrulewidth}
\multicolumn{7}{c}{Unsupervised} \\ 
\Xhline{1\arrayrulewidth} \\[-1.5ex]
SimCSE \cite{simcse}  & \textbf{72.9} & \textbf{84.0} & \textbf{75.6} & \textbf{84.8} & \textbf{81.8} & \textbf{79.8} \\ 
\texttt{cpt-text} S & 62.1 & 60.0 & 62.0 & 71.8 & 73.7 & 65.9 \\
\texttt{cpt-text} M & 62.7 & 62.8 & 64.6 & 73.9 & 75.3 & 67.9 \\
\texttt{cpt-text} L & 62.4 & 66.4 & 67.6 & 76.0 & 77.5 & 70.0 \\
\texttt{cpt-text} XL & 64.1 & 67.5 & 68.4 & 76.7 & 78.7 & 71.1 \\ 
\Xhline{2.5\arrayrulewidth}
\multicolumn{7}{c}{Transfer from NLI} \\ 
\Xhline{1\arrayrulewidth} \\[-1.5ex]
SimCSE \cite{simcse} & \textbf{77.5} & \textbf{87.3} & \textbf{82.4} & \textbf{86.7} & 83.9 & \textbf{83.6} \\
\texttt{cpt-text} S & 72.8 & 80.6 & 78.7 & 84.7 & 82.0 & 79.8 \\
\texttt{cpt-text} M & 73.7 & 80.2 & 78.9 & 85.0 & 82.8 & 80.1 \\
\texttt{cpt-text} L & 71.8 & 79.7 & 79.0 & 85.8 & 84.0 & 80.1 \\
\texttt{cpt-text} XL & 72.3 & 80.3 & 78.9 & 85.1 & \textbf{85.1} & 80.3 \\
\Xhline{2.5\arrayrulewidth}
\end{tabular}
\caption{\texttt{cpt-text} performs worse than the previous best sentence embedding method on sentence similarity tasks. We investigate this result in more detail in Section \ref{sec:behavior}.}
\label{table:sts}
\end{table}

\subsection{Text Search}

Previous work on training embedding methods for search typically requires fine-tuning on a particular text search dataset \cite{dpr, e2e, Qu}. It is also common to have a multi-step setup where fine-tuned models rely on an expensive query and document cross-attention encoder in the final step \cite{Qu, mini}. In contrast, we push the limits of using a \textit{single} embedding model for large-scale semantic search.

\subsubsection{Large-Scale Search}
\label{sec:large}

First, we evaluate our models on several large-scale text search benchmarks. MSMARCO \cite{msmarco} requires the model to search over 4M documents while Natural Questions (NQ) \cite{nq} and TriviaQA \cite{trivia} involve searching over 21M Wikipedia documents. We use the FAISS library \cite{Johnson} to build the vector indices for approximate $k$-nearest neighbor search. The \textit{same} unsupervised model discussed previously achieves impressive performance on semantic search. Table \ref{table:large} demonstrates that \texttt{cpt-text} outperforms prior unsupervised approaches by a big margin and larger model sizes consistently lead to improved performance. Surprisingly, on TriviaQA, our model is even competitive with fine-tuned models.

\begin{table}[]
\setlength{\tabcolsep}{4.5pt}
\centering
\begin{tabular}{lccc}
& MSMARCO  & NQ & TriviaQA  \\ 
\Xhline{2.5\arrayrulewidth}
Fine-tuned SOTA  & 44.3 & 84.8, 89.8  & 84.1, 87.8 \\
\Xhline{2.5\arrayrulewidth} \\[-1.5ex]
\multicolumn{4}{c}{Unsupervised} \\ 
\Xhline{1\arrayrulewidth} \\[-1.5ex]
BM25 & 18.4 & 62.9, 78.3 & 76.4, 83.2 \\
ICT & - & 50.9, 66.8 & 57.5, 73.6 \\
MSS & - & 59.8, 74.9 & 68.2, 79.4 \\
Contriever & - & 67.2, 81.3 & 74.2, 83.2 \\
\Xhline{1\arrayrulewidth} \\[-1.5ex]
\texttt{cpt-text} S & 19.9 & 65.5, 77.2 & 75.1, 81.7 \\
\texttt{cpt-text} M & 20.6 & 68.7, 79.6 & 78.0, 83.8 \\
\texttt{cpt-text} L & 21.5 & 73.0, 83.4 & 80.0, 86.8 \\
\texttt{cpt-text} XL & \textbf{22.7} & \textbf{78.8, 86.8} & \textbf{82.1, 86.9} \\
\Xhline{2.5\arrayrulewidth}        
\end{tabular}
\caption{Evaluation of unsupervised \texttt{cpt-text} models of different sizes on several large-scale text search benchmarks. We report MRR@10 on MSMARCO and Recall@20, Recall@100 for NQ and TriviaQA as done in prior work. Results for training with Inverse Cloze Task (ICT) and masked salient spans (MSS) objectives are taken from \citet{e2e}. \texttt{cpt-text} achieves the best results among unsupervised methods, surpassing keyword search methods on MSMARCO \cite{bm25} and embedding based methods \cite{contreiver} on NQ and TriviaQA.}
\label{table:large}
\end{table}

\subsubsection{BEIR Search}
\label{sec:beir}
Next, we evaluate our models on 11 zero-shot search tasks in the BEIR evaluation suite \cite{beir}. First, we observe that our unsupervised model performs competitively even with some previous embedding methods that leverage supervised MSMARCO data \cite{Xiong,tas}. Keyword-based BM25 \cite{bm25} achieves the best results in the unsupervised setting while \texttt{cpt-text} achieves the best transfer learning results.  

In the transfer setting, our models achieve a 5.2\% relative improvement over the previous best embedding method \cite{contreiver}. It also outperforms docT5query \cite{doct5} that relies on a fine-tuned T5 model \cite{t5} for document expansion. \texttt{cpt-text} results are competitive even with methods that use substantially more compute at test time. BM25+CE \cite{mini} uses keyword search to select top 100 documents which are then re-ranked by a cross-attention neural network encoder. The ranking encoder network performs computationally expensive joint query and document attention and cannot exploit indexing and approximate nearest neighbor algorithms for fast and efficient search at query time. Several other existing work take this approach of leveraging more computation resources at query time to obtain better search performance. ColBERT v2 \cite{colbert} is a multi-vector method that represents the query and the documents as a set of vectors, and employs a multi-step retrieval procedure to obtain relevant documents. Splade v2 \cite{splade} represents queries and documents as sparse vectors of size equivalent to the vocabulary of the BERT encoder \cite{bert}. Our \texttt{cpt-text} models compute only one dense embedding per document which are indexed offline and does not depend on any cross-attention re-ranker at query time.

\begin{table*}[]
\centering
\setlength{\tabcolsep}{5pt}
\setlength\tabcolsep{2.8pt} 
\begin{tabular}{lcccccccccccc}
& covid & nfc & fiqa & arg. & touche & quora & scifact & climate & dbp. & hotpot & fever & Avg. \\ 
\Xhline{2.5\arrayrulewidth}
\multicolumn{13}{c}{Unsupervised} \\ 
\Xhline{1\arrayrulewidth} \\[-1.5ex]
BM25 \cite{bm25} & \textbf{65.6} &	32.5 &	23.6 &	31.5 &	\textbf{36.7} &	78.9 &	66.5 &	\textbf{21.3} &	\textbf{31.3} &	\textbf{60.3} &	\textbf{75.3} &	\textbf{47.6} \\
Contriever \cite{contreiver} & 27.4 &	31.7 &	24.5 &	37.9 &	19.3 &	\textbf{83.5} &	64.9 &	15.5 &	29.2 &	48.1 &	68.2 &	40.9 \\  
\Xhline{1\arrayrulewidth} & \\[-1.5ex]
$\texttt{cpt-text}$ S & 52.9 & 32.0 & 34.1 &	38.7 &	21.0 &	68.1 &	65.4 &	15.8 &	27.2 &	51.5 &	57.1 &	42.2 \\
 $\texttt{cpt-text}$ M & 44.3 & 34.5 &	37.3 &	\textbf{41.2} &	23.3 &	70.3 &	68.3 &	15.6 &	29.6 &	53.0 &	58.2 &	43.2 \\
$\texttt{cpt-text}$ L & 42.7 &	\textbf{36.9} &	\textbf{39.7} &	39.2 &	22.8 &	68.7 &	\textbf{71.2} &	16.1 &	31.2 &	54.3 &	63.8 &	44.2 \\
\Xhline{2.5\arrayrulewidth}
\multicolumn{13}{c}{Transfer from MSMARCO} \\
\Xhline{1\arrayrulewidth} \\[-1.5ex]
TAS-B \cite{tas} & 48.1 &	31.9 &	30.0 &	42.9 &	16.2 &	83.5 &	64.3 &	22.8 &	38.4 &	58.4 &	70.0 &	46.0 \\
ANCE \cite{Xiong} & 65.4 &	23.7 &	29.5 &	41.5 &	24.0 &	85.2 &	50.7 &	19.8 &	28.1 &	45.6 &	66.9 &	43.7 \\
Contriever \cite{contreiver}  & 59.6 &	32.8 &	32.9 &	44.6 &	23.0 &	\textbf{86.5} &	67.7 &	23.7 &	41.3 &	63.8 &	75.8 &	50.2 \\ 
\Xhline{1\arrayrulewidth} \\[-1.5ex]
$\texttt{cpt-text}$ S & 67.9 &	33.2 &	38.4 &	47.0 &	28.5 &	70.6 &	67.2 &	18.5 &	36.2 &	59.4 &	72.1 & 49.0 \\
$\texttt{cpt-text}$ M & 58.5 &	36.7 &	42.2 &	\textbf{49.2} &	29.7 &	69.7 &	70.4 &	19.9 &	38.6 &	63.1 &	77.0 &	50.5    \\
$\texttt{cpt-text}$ L & 56.2 &	38.0 &	45.2 &	46.9 &	30.9 &	67.7 &	74.4 &	19.4 &	41.2 &	64.8 &	75.6 &	50.9    \\
$\texttt{cpt-text}$ XL & 64.9 & \textbf{40.7} &	\textbf{51.2} &	43.5 &	29.1 &	63.8 &	\textbf{75.4} & 22.3 & 43.2 & \textbf{68.8} & 77.5 & \textbf{52.8} \\
\Xhline{1\arrayrulewidth} & \\[-1.5ex]
docT5query \cite{doct5} & 71.3 & 32.8 & 29.1 & 34.9 & \textbf{34.7} & 80.2 & 67.5 & 20.1 & 33.1 & 58.0 & 71.4 & 48.5 \\
BM25+CE \cite{mini} & \textbf{75.7} & 35.0 & 34.7 & 31.1 & 27.1 & 82.5 & 68.8 & \textbf{25.3} & 39.2 & 70.7 & 81.9 & 52.0 \\
ColBERT v2 \cite{colbert} & 73.8 & 33.8 & 35.6 & 46.3 & 26.3 & 85.2 & 69.3 & 17.6 & \textbf{44.6} & 66.7 & 78.5 & 52.5 \\
Splade v2 \cite{splade} & 71.0 & 33.4 & 33.6 & 47.9 & 27.2 & 83.8 & 69.3 & 23.5 & 43.5 & 68.4 & \textbf{78.6} & 52.7 \\
\Xhline{2.5\arrayrulewidth}
\end{tabular}
\caption{Comparison of $\texttt{cpt-text}$ to previous methods on 11 zero-shot search tasks in the BEIR evaluation suite \cite{beir}. Results are reported both in the unsupervised data setting and in the transfer data setting. $\texttt{cpt-text}$ outperforms previous best embedding methods \cite{Xiong,tas,contreiver} in both the settings. In the unsupervised setting, BM25 \cite{bm25} still achieves the best performance while in the transfer setting $\texttt{cpt-text}$ is competitive with methods that use substantially more compute at test time \cite{mini,colbert,splade}.}
\label{table:beir}
\end{table*}

\subsection{Code Search}
\label{sec:code}

We evaluate our code embedding models on the code search task using the CodeSearchNet benchmark \cite{codesearchnet}. Given a natural language query, the model is expected to retrieve the relevant code block among 1K candidates. The models are evaluated on 6 programming languages and our model achieves state-of-the-art results (Table \ref{table:code}). Unlike with text embeddings, we do not see a performance improvement with increased model size for code embeddings.

\begin{table}[]
\small
\setlength{\tabcolsep}{4pt}
\centering
\tabcolsep=0.105cm
\begin{tabular}{lccccccc}
& Go & Ruby & Python & Java & JS & PHP & Avg. \\
\Xhline{2.5\arrayrulewidth} \\[-1.5ex]
\small{CodeBERT} & 69.3 & 70.6 & 84.0 & 86.8 & 74.8 & 70.6 & 76.0 \\
\small{GraphCodeBERT} & 84.1 & 73.2 & 87.9 & 75.7 & 71.1 & 72.5 & 77.4 \\
\Xhline{1\arrayrulewidth}  \\[-1.5ex]
\texttt{cpt-code} S & \textbf{97.7} & \textbf{86.3} & 99.8 & 94.0 & 86.0 & 96.7 & 93.4 \\
\texttt{cpt-code} M & 97.5 & 85.5 & \textbf{99.9} & \textbf{94.4} & \textbf{86.5} & \textbf{97.2} & \textbf{93.5} \\
\Xhline{2.5\arrayrulewidth}
\end{tabular}
\caption{Comparison of $\texttt{cpt-code}$ on code search across 6 programming languages \cite{codesearchnet} with CodeBERT \cite{codebert} and GraphCodeBERT \cite{Guo}. The task requires finding the relevant code block among 1K candidates for a given natural language query. $\texttt{cpt-code}$ performs substantially better than previous methods on all the languages.}
\label{table:code}
\end{table}

We also evaluate on a harder setting of finding the relevant code block among 10K candidates instead of 1K. Here, we compare the performance of \texttt{cpt-text} models against \texttt{cpt-code} models (Table \ref{table:large_code}). It is interesting to see that text embedding performs fairly well in code search especially in Python. We see a drop in performance for code embedding models with increased distractors and still don’t see bigger models giving a boost in search performance. 

\begin{table}[]
\small
\setlength{\tabcolsep}{4pt}
\centering
\tabcolsep=0.11cm
\begin{tabular}{lccccccc}
 & Go & Ruby & Python & Java & JS & PHP & Avg. \\ 
\Xhline{2.5\arrayrulewidth} \\[-1.5ex]
\texttt{cpt-text} S & 60.6 & 58.9 & 92.6 & 48.4 & 52.8 & 47.6 & 60.1 \\
\texttt{cpt-text} M & 65.4 & 63.1 & 91.4 & 47.9 & 53.5 & 43.1 & 60.7 \\ 
\Xhline{1\arrayrulewidth} & \\[-1.5ex]
\texttt{cpt-code} S & \textbf{90.4} & 80.6 & 98.8 & \textbf{81.9} & \textbf{76.1} & \textbf{85.3} & \textbf{85.5} \\
\texttt{cpt-code} M & 90.0 & \textbf{89.1} & \textbf{98.9} & 81.1 & 75.6 & 85.1 & 85.0 \\
\Xhline{2.5\arrayrulewidth}
\end{tabular}
\caption{Comparison of \texttt{cpt-code} vs \texttt{cpt-text} on large scale code search \cite{codesearchnet}. The task is to retrieve the relevant code block among 10K candidates for a given natural language query. It is interesting to note that $\texttt{cpt-text}$ performs quite well on Python code search without explicitly training on (text, code) pairs.}
\label{table:large_code}
\end{table}

\subsection{Analysis}

\subsubsection{Effect of Batch Size}
\label{sec:bs}
Our ablation study highlights the effect of the model's batch size on the final performance. Table \ref{table:bs_effect} compares the performance of S (300M) \texttt{cpt-text} model trained with different batch sizes on the NQ development set. Since we train with in-batch negatives, a larger batch increases the chances of having hard negatives in a batch, resulting in a significant performance boost.

\begin{table}[]
\centering
\begin{tabular}{l|c}
\Xhline{2.5\arrayrulewidth}
Batch Size & MRR@10 \\ 
\hline  
1536       & 71.4   \\ 
12288      & 84.7   \\ 
\Xhline{2.5\arrayrulewidth}
\end{tabular}
\caption{Performance of the \texttt{cpt-text} 300M model on NQ dev set given different training batch sizes.}
\label{table:bs_effect}
\end{table}

\subsubsection{Training Behavior}
\label{sec:behavior}

\begin{figure}[h]
\centering
\includegraphics[width=8cm]{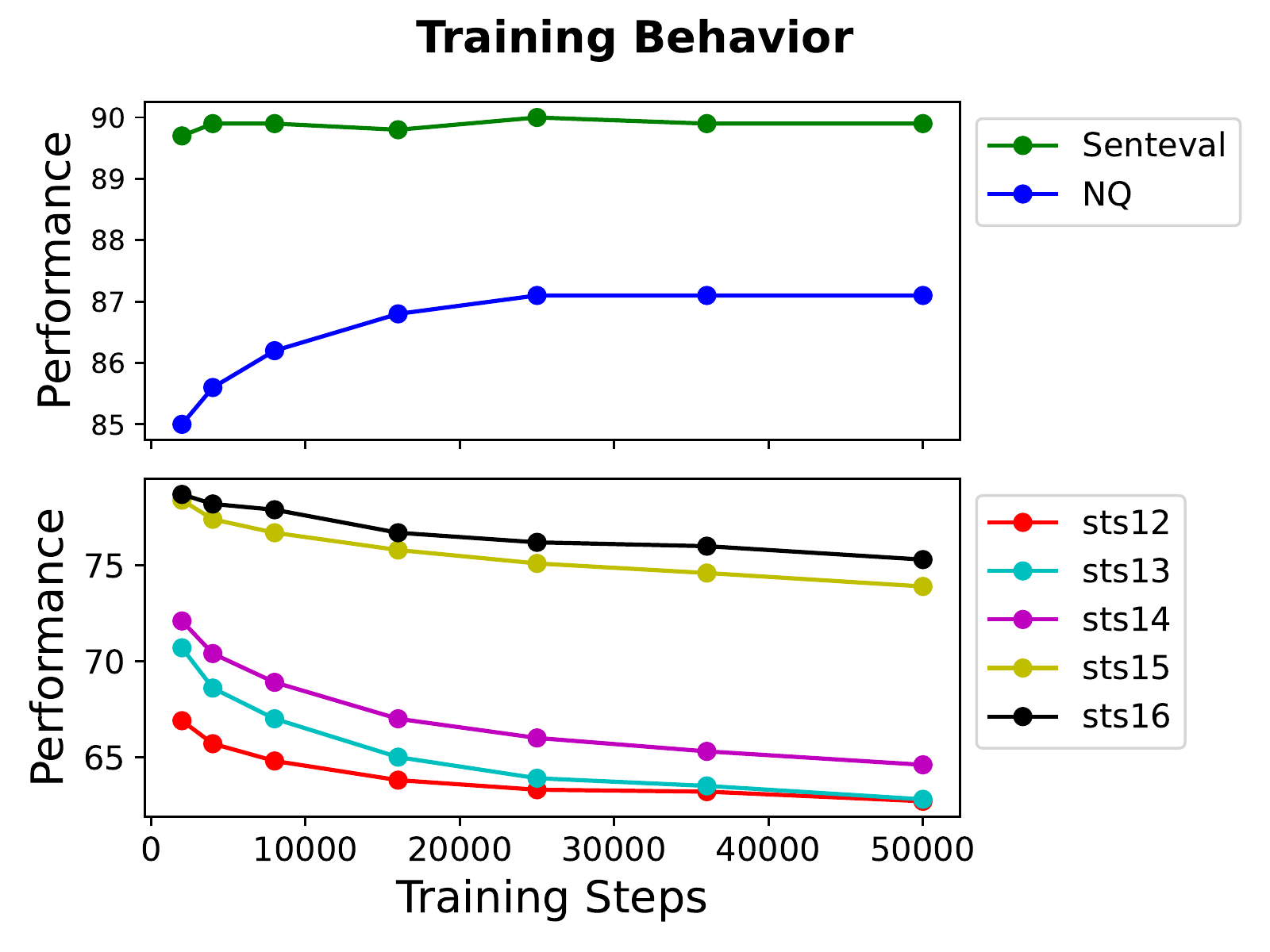}
\caption{Performance of M (1.2B) $\texttt{cpt-text}$ model on classification, search and sentence similarity tasks at different training steps. While the performance on search and classification improves with longer training, the performance on sentence similarity degrades.}
\label{fig:behavior}
\end{figure}

We observe that as we train our models for longer, the performance on search and classification tasks increases while the performance on sentence similarity tasks decreases (Figure \ref{fig:behavior}). As discussed previously, sentence similarity is not a well defined task. A hypothesis is that search tasks and sentence similarity tasks might have contradicting definitions. For example, a sentence and its negation could be considered as relevant during search, but  not ``similar" in sentence similarity tasks. It is also important to note that previous embedding search methods do not report performance on sentence similarity tasks \cite{dpr, e2e, contreiver} and previous sentence embedding methods do not evaluate on search tasks \cite{sbert,declutr,simcse}. When deciding the model checkpoints to use for evaluation, we assigned higher importance to search and classification tasks as they are commonly associated with clearly defined real-world applications while sentence similarity tasks are less so. 


\section{Related Work}
The goal of representation learning \cite{bengio} is to learn an embedding space in which similar examples stay close to each other while dissimilar ones are far apart \cite{hadsell}. In contrastive learning, the learning procedure is formulated as a classification problem given similar and dissimilar candidates \cite{chopra,Gutmann,Schroff,inbatch,Oord}. Recent work relies on contrastive objective to learn representations for images \cite{wu,He,Chen,Zbontar}, text, or both jointly \cite{Lu,Sun,Kim,clip,Khosla}. In self-supervised contrastive learning, positive samples can be collected in various approaches including by creating an augmented version of the original input without modifying the semantic meaning \cite{simcse}, by grouping samples within the same context \cite{declutr,contreiver}, or by collecting data about the same object from different views \cite{tian}. 

Learning word embeddings is a well studied research area \cite{brown,Gutmann,mikolov,pennington}. Learning low-dimensional representations of larger text pieces, denser than raw term-based vectors, has been studied extensively as well \cite{deerwester,contrastive}. Most of the recent models for learning sentence embeddings rely on supervised NLI datasets, using entailment pairs as positive examples and contradiction pairs as (hard) negatives. SBERT \cite{sbert} trained a siamese network to learn a representation where sentence similarity is estimated by the cosine similarity between embeddings. \citet{Li} improves the embedding space to be isotropic via normalizing flows. The whitening operation is another alternative operation to improve the isotropy of the embedding space \cite{Su}. It is typical to initialize such models with a pre-trained language model \cite{bert} before training on NLI datasets.

Several methods have been studied for unsupervised or self-supervised sentence embedding learning \cite{Logeswaran,Zhang,simcse}. Common approaches consider sentences within the same context as semantically similar samples \cite{Kiros,Logeswaran}. To create positive training pairs with augmented samples, a diverse set of text augmentation operations have been explored, including lexicon-based distortion \cite{Wei}, synonym replacement \cite{Kobayashi}, back-translation \cite{Fang}, cut-off \cite{Shen} and dropout \cite{simcse}. However, unsupervised sentence embedding models still perform notably worse than supervised sentence encoders.

Large-scale text search based on dense embeddings and neural information retrieval (neural IR) have the potential to generalize better than keyword matching in classic IR systems. Neural IR systems encode documents at the indexing stage and then perform nearest neighbor search \cite{Johnson} at query time \cite{Lin}. Neural IR models are usually learned by fine-tuning a pre-trained language model on supervised search corpus \cite{ORQA,REALM,Karpukhin,Lewis}. Many SOTA search models combine classical IR with neural IR in a staged setup, where the candidates are first narrowed down by BM25 keyword search \cite{bm25} and then re-ranked by joint query and document neural encoders \cite{Nogueira,Qu}. \citet{Xiong} proposed ANCE, a contrastive learning framework for learning text representations for dense retrieval using mined hard negatives.  Other unsupervised retriever methods use the Inverse Cloze Task or masked salient spans to achieve significant improvement on ODQA tasks \cite{e2e}. In comparison to most prior work, we find that with a large enough batch size, it is possible to achieve good search performance without using supervised data. Finally, the recently published Contriever \cite{contreiver} is most similar to our work on learning text embeddings for text search using contrastive learning on unlabeled data.  

Semantic code search refers to the task of retrieving code relevant to a query in natural language. The CodeSearchNet challenge \cite{codesearchnet} presents a set of benchmark code search tasks in different programming languages, as well as a simple baseline model to predict embeddings of query and code via contrastive learning on a dataset of (text, code) pairs. ContraCode  \cite{Jain} uses a contrastive learning task of identifying functionally similar programs, where the functionally similar samples are generated via source-to-source compiler transformations. CodeBERT \cite{codebert} learns to predict semantic similarity with a pre-trained language model and GraphCodeBERT \cite{Guo} further improves the performance on the CodeSearchNet benchmark by adding pre-training tasks on code structure. 

\section{Broader Impacts}
Prior research has shown that text representation models encode the biases present in their training data, including those which are discriminatory towards protected groups such as Black people or women \cite{Bolukbasi,Caliskan, May,Zhao,Rudinger}. Biases encoded in embedding models may cause representational harms\footnote{ Representational harms occur when systems reinforce the subordination of some groups along the lines of identity, e.g. stereotyping or denigration \cite{Crawford}.} by reinforcing existent societal biases in the text corpus, and further propagating them in downstream tasks of embedding models. 

Therefore, we encourage further research on two research agendas: (a) developing robust evaluation methodologies for multiple classes of bias in training data and pre-trained models, and (b) developing and improving methods for mitigating encoded bias, including fine-tuning to reduce bias in pre-trained models \cite{Caliskan,May,Bolukbasi,Liang,parkbias, Solaiman}. Until we have robust evaluation methodology, it is important to restrict and monitor the use of the model in downstream applications. Particularly for those where risk of representational harm is great and those where biased representations may influence the allocation of resources and opportunities to people.

Our embedding models are trained with large batch sizes and require substantial computation resources. While this training regime is environmentally and computationally costly, there are promising paths forward to amortize and offset these costs while allowing users to benefits from the capabilities of these models. For example, safe public access to large pre-trained language models, and efficient training pipelines that leverage improved model architectures and training schemes. We encourage further research and implementation efforts in these areas.


\section{Conclusion}
We showed that contrastive pre-training on unsupervised data with a sufficiently large batch size can lead to high quality vector representations of text and code. Our models achieved new state-of-the-art results in linear-probe classification, text search and code search. We find that our models underperformed on sentence similarity tasks and observed unexpected training behavior with respect to these tasks.  Finally, we discussed the broader impact of our work on society. 
\bibliography{main}
\bibliographystyle{icml2022}

\end{document}